\documentclass[
hf,
]{ceurart}

\usepackage{listings}
\usepackage{subfiles}
\usepackage{footmisc}
\usepackage{todonotes}
\usepackage[inline]{enumitem}
\usepackage[normalem]{ulem}
\lstdefinestyle{mystyle}{
    basicstyle=\ttfamily\footnotesize,
    breakatwhitespace=false,         
    breaklines=true,                 
    captionpos=b,                    
    keepspaces=true,                 
    showspaces=false,                
    showstringspaces=false,
    showtabs=false,                  
    tabsize=2
}

\begin{document}

%%
%% Rights management information.
%% CC-BY is default license.
\copyrightyear{2022}
\copyrightclause{Copyright for this paper by its authors.
  Use permitted under Creative Commons License Attribution 4.0
  International (CC BY 4.0).}

%%
%% This command is for the conference information
\conference{WOP2022: 13th Workshop on Ontology Design and Patterns,
  October 23--24, 2022, Hangzhou, China}

%%
%% The "title" command
\title{The Music Annotation Pattern}

%%
%% The "author" command and its associated commands are used to define
%% the authors and their affiliations.
\author[1]{Jacopo de Berardinis}[%
orcid=0000-0001-6770-1969,
email=jacopo.deberardinis@kcl.ac.uk,
]
\cormark[1]
%\fnmark[1]

\author[1]{Albert Mero\~{n}o-Pe\~{n}uela}[%
orcid=0000-0003-4646-5842,
email=albert.merono@kcl.ac.uk,
]
\address[1]{King's College London}
%\fnmark[1]

\author[2]{Andrea Poltronieri}[%
orcid=0000-0003-3848-7574,
email=andrea.poltronieri2@unibo.it,
]
\cormark[1]
%\fnmark[1]

\author[2]{Valentina Presutti}[%
orcid=0000-0002-9380-5160,
email=valentina.presutti@unibo.it,
]
\address[2]{University of Bologna}
%\fnmark[1]

%% Footnotes
\cortext[2]{Corresponding author.}
%\fntext[1]{These authors contributed equally.}

%%
%% The abstract is a short summary of the work to be presented in the
%% article.
\iffalse
\begin{abstract}
  Music material is notoriously insidious to represent because of its multifacetedness. To cope with this complexity, numerous systems for representing music have been developed as independent standards over the past decades, but without taking into account the problem of interoperability.
  Moreover, most of these systems lack the essential semantic expressiveness to represent the complex relationships of the musical language and struggle with different types of annotations, such as annotations extracted from audio and score.
  In this article, we present the \emph{Music Notation Pattern}, an Ontology Design Pattern that allows to represent different types of musical annotations by taking into account the semantics of the elements contained in them.
  Moreover, the proposed pattern allows to describe annotations derived from different sources, such as audio and score, and according to heterogeneous timing conventions.
\end{abstract}
\fi

\begin{abstract}
  The annotation of music content is a complex process to represent due to its inherent multifaceted, subjectivity, and interdisciplinary nature.
  Numerous systems and conventions for annotating music have been developed as independent standards over the past decades.
  Little has been done to make them interoperable, which jeopardises cross-corpora studies as it requires users to familiarise with a multitude of conventions.
  Most of these systems lack the semantic expressiveness needed to represent the complexity of the musical language and cannot model multi-modal annotations originating from audio and symbolic sources.
  In this article, we introduce the \emph{Music Annotation Pattern}, an Ontology Design Pattern (ODP) to homogenise different annotation systems and to represent several types of musical objects (e.g. chords, patterns, structures). This ODP preserves the semantics of the object's content at different levels and temporal granularity.
  Moreover, our ODP accounts for multi-modality upfront, to describe annotations derived from different sources, and it is the first to enable the integration of music datasets at a large scale.
\end{abstract}

%%
%% Keywords. The author(s) should pick words that accurately describe
%% the work being presented. Separate the keywords with commas.
\begin{keywords}
  Semantic Web \sep
  Ontology \sep
  Music Information Retrieval \sep
  Computational Musicology
\end{keywords}

%%
%% This command processes the author and affiliation and title
%% information and builds the first part of the formatted document.
\maketitle

\section{Introduction}\label{sec:introduction}

Similarly to other forms of artistic expression, the analysis of music can be considered as a quest for meaning -- a process driven by musical theories and perceptual cues attempting to shed light on the potentially ambiguous and intricate messages that artists have encoded in their music \cite{pople2006theory}.
Starting from a composition or a performance, music analysis usually focuses on detecting elements related to harmony, form, texture, etc., along with the identification of potential interrelated functions they may exert in the piece (creating or releasing tension, evoking images, inducing emotions, etc.) \cite{johnston2009harmony}.

At the core of this multifaceted process lies the ability to effectively annotate music.
%\sout{encode the output of the analyses as \textit{music annotations}} %\footnote{In this work, we refer to music annotations as the tangible outcomes of the annotation process resulting of the music analysis.}
For example, if the goal of a harmonic analysis is to identify chords from a composition, a music annotation may correspond to a list of chords together with a reference to their onset (i.e. when they occur in the piece).
Besides contributing to the more general goal of understanding music, these annotations are also of pedagogic interest (e.g. teaching material for classrooms in analysis, harmony, or composition) and of musicological relevance.
They also provide valuable data for training and evaluating algorithmic methods for music information retrieval (MIR) and computational music analysis (CMA), and for supporting performers studying scores and preparing their own interpretation \cite{giraud2018dezrann}.
This interdisciplinary interest in music annotations has also fuelled the development of novel applications and workflows focused on their collection \cite{turnbull2007game}, interaction \cite{pugin2018interaction} and sharing \cite{giraud2018dezrann}.

Nevertheless, annotating music has always been a challenging task in many respects \cite{hadjakos2017challenges}.
Musical content is rich in components (voices, sections, etc.) and nuances (accents, prolongations, modulations) that are often difficult to represent and to consistently relate to the content of an annotation.
Several types of musical notations have been introduced to address this problem, although primarily focused on representing musical scores (c.f. Section~\ref{ssec:modelling-scores}).
Even the score itself is based on conventions and symbols that have evolved diachronically -- as musical periods have changed, as well as stylistically -- as musical genres vary \cite{savage2019cultural}.
This evanescent aspect of music is then more pronounced when focusing on representing music annotations.
For example, when annotating chords, different notation systems have been used over the years, starting with the \emph{basso continuo}, almost universally used in the Baroque era, to the modern \emph{Leadsheet notations}, mainly used to annotate chords in Jazz music \cite{kite2012performer}.
% These types of notation are used for different purposes, and they also reflect the changing concept of chord.

%Despite the transition of music and musicology towards digital tools, the problem of how to annotate music still requires more research efforts.
A multitude of notation systems have been developed, proposing different approaches on how to annotate music.
This fragmentation is reflected in a vast heterogeneity of file formats and extensions, with consequent interoperability problems.
When annotations are encoded within a score, software tools for music processing and computer-aided musicology, like \texttt{music21} \cite{cuthbert2010music21} and \textit{note-seq}\footnote{\url{https://github.com/magenta/note-seq}}, have rapidly evolved to parse a variety of symbolic formats \footnote{See, for example \url{https://web.mit.edu/music21/doc/moduleReference/moduleConverter.html}}.
When annotations are decoupled from the music content \cite{vsevolod_eremenko_2018_1290737, neuwirth2018annotated}, these are often encoded using dataset-specific standards and conventions.
As a result, retrieving and integrating music annotations from different sources is a challenging, time-consuming task, which stems from the encoding problem and the lack of well-established standards for releasing music datasets \cite{carriero2021semanticintegration}.
This brings a cascade of effects: 
\begin{enumerate*}[label=(\roman*)]
  \item it limits the ability to perform cross-corpora studies, especially in multi-modal settings -- involving both audio and score annotations;
  \item it leaks ambiguity in the annotations due to the poor semantic expressiveness of the current approaches; and
  \item it confines users to familiarising with a multitude of standards.
\end{enumerate*}

In this article, 
%we present the JAMS Ontology, an ontology for modelling semantically enriched chord annotations. 
%The JAMS Ontology is part of a network of Ontolgies modelled in the context of the Polifonia Project \cite{carriero2021semanticintegration}.
%Moreover, together the JAMS Ontology, 
we introduce the \emph{Music Annotation Pattern}, an Ontology Design Pattern for modelling a wide set of music annotations. 
The \emph{Music Annotation Pattern} is a reusable block for representing annotations of different types, from different sources, and addressing  heterogeneous timing conventions.
The ODP has been used in preliminary experiments integrating harmonic datasets (chord annotations from multiple sources) in the Polifonia project\footnote{\label{polifonia}\url{https://polifonia-project.eu}}.
To our best, it is the first attempt at achieving semantic interoperability of music annotations collected from multi-modal sources.

\section{Related Work}\label{sec:related}

% One of the greatest challenges in the study of music is undoubtedly related to the representation of musical content.
The complexity of representing musical content is related to the manifold sources that are available when studying music.
% To this extent, 
To contextualise this process, Vinet \cite{vinet2003representation} introduces two different \emph{Representation Levels} to categorise different types of music content: \emph{signal representations} and \emph{symbolic representations}.
A symbolic representation is context-aware and describes events in relation to formalised concepts of music (music theory), whereas the signal representation is a blind, context-unaware representation, thus adapted to transmit any, non-musical kind of sound, and even non-audible signals.

%Wiggins et al. in 1993 proposed a system based on two orthogonal dimensions, on which to evaluate different music representation systems, namely \emph{expressive completeness} and \emph{structural generality}. 
%According to such categories, audio signal representations revealed to be very expressive but very little structured. 
%On the contrary, symbolic representations are by definition structured, but most of the times not very expressive. 

In this paper, we focus on symbolic representation systems and how these can be semantically described to address the three challenges outlined in the introduction.\footnote{This does not  imply that a symbolic annotation cannot also refer to audio music (\textit{alias} tracks, recordings).}

\subsection{Modelling scores and score-embedded annotations}\label{ssec:modelling-scores}

Over the years, various representation systems have been developed, some of which are still used today. %, while others are no longer employed.
A notable example is MIDI (Musical Instrument Digital Interface) \cite{international1983midi}, which also provides a data communication protocol for music production and live performance.
%MIDI, however, is a data communication protocol, i.e. an agreement among manufacturers of music equipment, computers, and software that describes a means for music systems for exchanging information and control signals .
A MIDI file can be described as a stream of events, each defined by two components: \emph{MIDI time} and \emph{MIDI message}.
The time value describes the time to wait (a temporal offset) before executing the following message. The message value, instead, is a sequence of bytes, where the first one is a command, often followed by complementary data.
% The command byte determines the type of command, and incorporates information about the channel the event is associated with. The available channels are 16 (from 0 to 15), corresponding to 16 independent instruments. 
% The following data bytes can express a discrete range of properties of the MIDI file, such as pitch and velocity.

The ABC notation \cite{walshaw2011} is a text-based music notation system and the de facto standard for folk and traditional music. 
An ABC tune consists of a \emph{tune header} and a \emph{tune body}, terminated by an empty line or the end of the file.
The \emph{tune header} contains the tune's metadata, and can be filled with 27 different fields that describe composer, tempo, rhythm, source, etc.
The tune body, instead, describes the actual music content, such as notes, rests, bars, chords, and clefs.

MusicXML \cite{good2001musicxml} is an XML-based music interchange language.
It is intended to represent common western musical notation from the seventeenth century onwards, including both classical and popular music. Similarly to MIDI, MusicXML defines both an interchange language and a file format (in this case XML).

The Music Encoding Initiative (MEI) \cite{roland2002music} is a community-driven, open-source effort to define a system for encoding musical documents in a machine-readable structure. The community formalised the MEI schema, a core set of rules for recording physical and intellectual characteristics of music notation documents expressed with an XML schema. This framework aims at preserving the XML compatibility while expressing a wide level of music nuances.

Other systems of symbolic notation  include the \emph{CHARM system} \cite{smaill1993charm}, \emph{**kern} \cite{humdrum2002} and \emph{LilyPond} \cite{nienhuys2003lilypond}.
All these formats differ dramatically in their syntax, % of these notation systems is very different and the software used to process these file types does not always support all formats. 
which may exacerbate the interoperability problem and the consequent fragmentation of music data.

\subsection{Modelling decoupled annotations}

To overcome these problems, annotation standards have been proposed to decouple annotations from the scores, and to encode them in a separate yet unified format.
The most notable example is the Annotated Music Specification for Reproducible MIR Research (JAMS) \cite{humphrey2014jams, mcfee2015jams}, a JSON-based format to encode music annotations. It is primarily used to train and evaluate MIR algorithms, especially in the audio domain.
JAMS supports the annotation of several music object types -- from notes and chords to patterns and emotions, unambiguously defining the onset, duration, value and confidence of each observation (e.g. "C:major" starting at second 3, lasting for 4 seconds, detected with a confidence level of $90\%$). This standard also offers the possibility of storing multiple and heterogeneous annotations in the same file, as long as they pertain to the same piece. Notably, JAMS provides a loose schema to record metadata, both related to the track (title, artists, etc.) and to each annotation (annotator, annotation tools, etc.).

Nonetheless, JAMS supports annotations collected from signal representation (audio), as it was not originally designed for the symbolic domain.
This is due to a discrepancy between audio-based annotations -- expressing temporal information in absolute times (seconds), and symbolic annotations -- using relative or metrical temporal anchors (e.g. beats, measures).
Also, from a descriptive perspective, it is not possible to disambiguate certain attributes in the metadata sections. For instance, the ``\texttt{artist}'' field in the current JAMS definition may refer to the composer or to the performer of the piece.
Finally, JAMS is limited to the expressiveness of JSON, which does not allow for the semantic expression of concepts that are sometimes essential for describing musical content.
For example, even if the specification of composers and performers was possible in the standard, this would still be insufficient to express the semantic relationships occurring between these concepts.

\subsection{Modelling semantics in music data}

To encode semantics in music data, and account for the ambiguity problem in music annotations, Semantic Web technologies can be useful, as shown in other domains such as Cultural Heritage \cite{carriero2019arco}.
Over the past two decades, several ontologies have been developed in the music domain.
Some ontologies have been designed for describing high-level descriptive and cataloguing information, such as the The Music Ontology \cite{raimond2007} and the DOREMUS Ontology \cite{lisena2017}.

Other ontologies describe musical notation, both from the music score and the symbolic points of view. For example, the MIDI Linked Data Cloud \cite{merono2017} models symbolic music descriptions encoded in MIDI format.
The Music Theory Ontology (MTO) \cite{rashid2018} aims to describe theoretical concepts related to a music composition, while The Music Score Ontology (Music OWL) \cite{jones2017} represents similar concepts with a focus on music sheet notation. Finally, the Music Notation Ontology \cite{cherfi2017} focuses on the core ``semantic'' information present in a score.
%Other ontologies aim to describe specific aspects of the musical domain, such as the Chord Ontology, the Tonality Ontology, the Temperament Ontology \cite{fazekas2010}, and the Segment Ontology \cite{fields2011}.
The Music Encoding and Linked Data framework (MELD) \cite{page2019meld} reuses multiple ontologies, such as the Music and Segment Ontologies, FRBR in order to describe real-time annotation of digital music scores.
The Music Note Ontology \cite{poltronieri2021musicnoteontology} proposes to model the relationships between a symbolic representation and the audio representation, but only considering the structure of the music score and the granularity level of the music note.

%In sum, none of these ontologies provides a comprehensive, scalable, and coherent representation of music notation. 
Each of these ontologies covers a specific aspect of music notations. 
Our ODP reuses and extend their modelling solutions  to provide a comprehensive, scalable and coherent representation music annotations.
%To the best of our knowledge, there is no ontology to date capable of describing a piece of music and the different types of notations contained in it.  

%\section{The JAMS Ontology}

%\subfile{sections/3_JAMS_ontology}

\section{The Music Annotation Pattern}\label{sec:pattern}

% The JAMS Ontology is, however, limited to the description of chord annotations only. 
%To address the aforementioned problems, we propose the \emph{Music Annotation Pattern}.
The Music Annotation ODP addresses the goal of modelling different types of musical annotations.
%has the objective of offering a model in the field of ontology engineering to be reused for modelling different types of musical annotations. 
For example, this ODP can be used to describe musical chords, notes rather than patterns, both harmonic and melodic and structural annotations. 
The \emph{Music Annotation} ODP also aims to represent annotations derived from different types of sources, such as audio and score. 

\begin{figure*}[b]
    \centering
    \includegraphics[width=\textwidth]{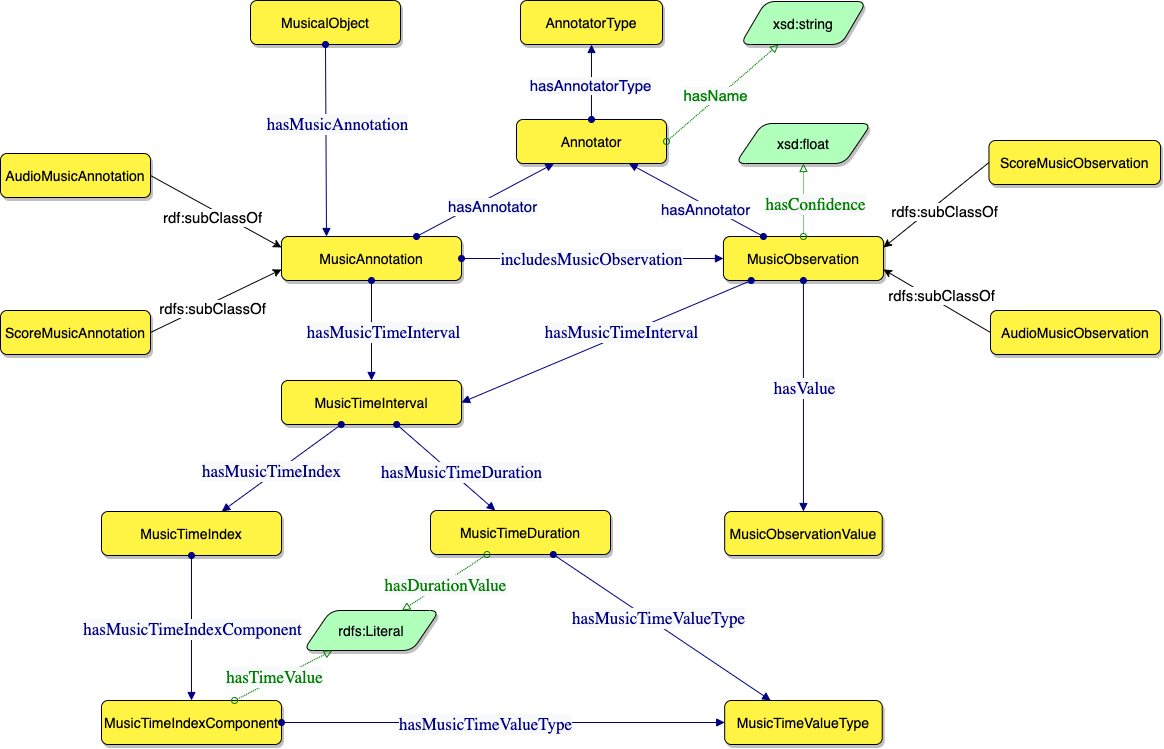}
    \caption{The Music Annotation Pattern. We use the Graffoo notation: yellow boxes are classes, blue/green arrows are object/datatype properties, purple circles are individuals, green polygons are datatypes.}
    \label{fig:annotation_pattern}
\end{figure*}

\noindent The ODP is represented in Figure \ref{fig:annotation_pattern} an it is available online at the following URI:

\begin{center}
\texttt{\url{https://purl.org/andreapoltronieri/music-annotation-pattern}}
\end{center}

\noindent The complete implementation and documentation of the pattern, as well as its documentation and all the examples presented in this paper are available on a dedicated GitHub repository\footnote{\label{github}Music Annotation Pattern repository: \url{https://github.com/andreamust/music-annotation-pattern}}.

\noindent To be compliant with the practice of the Music Information Retrieval community, we reuse the terminology from JAMS\footnote{Official JAMS documentation: \url{https://jams.readthedocs.io/en/stable/}} \cite{humphrey2014jams}. 
The following terms are used for the ODP vocabulary:
\begin{itemize}
    \item \emph{Music Annotation: } a music annotation is defined as a group of \texttt{Music Observations} (see below) that share certain elements, such as the method used for the annotation and the type of object being annotated (e.g. chords, notes, patterns); an annotation has one and only one annotator, that can be of different types e.g., a human, a computational method, and which is the same for all its observations.
    \item \emph{Music Observation: } a music observation is defined as the content of a music annotation. It includes all the elements that characterise the observation. For example, in the case of an annotation of chords, each observation is associated with one chord, and it specifies, in addition to the chord value, its related temporal information and the confidence of the annotator for that observation. 
\end{itemize}
\noindent
The structure of the Music Annotation ODP consists of the relations between an \texttt{MusicAnnotation} and its \texttt{MusicObservations}.

%The JAMS ontology was modelled and tested following the approach named eXtreme design \cite{presutti2009xd}.
\noindent
An integration effort of a set of datasets containing chord annotations, in the context of the Polifonia project\footref{polifonia}, provided a useful empirical ground to define a set of Competency questions (CQs) to drive the design of the Music Annotation ODP. 
They are listed in Table \ref{tab:pattern-cqs}. Each competency question is associated with a corresponding SPARQL query, they are all available on the project's GitHub repository\footref{github}. 

\noindent
The ODP was modelled by following a CQ-driven approach~\cite{presutti2009xd}, and by reusing a JAMS-based terminology. 
%You can find all SPARQL queries in the project's GitHub repository\footref{github}. 

\begin{table}[ht]
\begin{tabular}{ll}
\hline
ID  & Competency questions                                    \\
\hline
CQ1 & What is the type of a music annotation/observation for a musical object?        \\
CQ2 & What is the time frame within the musical object addressed by an annotation?                   \\
CQ3 & What is its start time (i.e. the starting time of the time frame)?                        \\
CQ4 & Which are the observations included in an annotation?         \\
CQ5 & \makecell[l]{For a specific music observation, what is the starting point of its addressed time frame, within \\ its reference musical object?} \\
CQ6 & For a specific music observation, what is its addressed time frame, within the musical object?      \\
CQ7 & What is the value of a music observation?  \\
CQ8 & Who/what is the annotator of a music annotation/observation, and what is its type?  \\
CQ9 & What is the confidence of a music observation?  \\
CQ10 & What is the musical object addressed by a music annotation?  \\
\hline
\end{tabular}
\caption{Competency questions addressed by the Music Annotation ODP.}
\label{tab:pattern-cqs}
\end{table}

\paragraph{Annotation.}
Addressing CQ1, CQ4, CQ10: a \texttt{MusicAnnotation} has to be intended as a collection of \texttt{MusicObservation}s about a \texttt{MusicalObject}. For musical objects, in this context, we refer to a concept generalising over audio tracks and scores. 
\texttt{MusicAnnotations} can be of two types: \texttt{ScoreMusicAnnotation} and \texttt{AudioMusicAnnotation}.

%\begin{lstlisting}
%Class: MusicAnnotation
%    SubClassOf: 
%        hasMusicTimeInterval some MusicTimeInterval,
%        hasAnnotator only Annotator,
%        includesMusicObservation only MusicObservation,
%        isMusicAnnotationOf only MusicalObject,
%        includesMusicObservation min 1 MusicObservation,
%        hasMusicTimeInterval exactly 1 MusicTimeInterval,
%        isMusicAnnotationOf exactly 1 MusicalObject
%\end{lstlisting}

% These two types of annotation define the encoding to be used for representing time information. 

\paragraph{Time information.}
Addressing CQ2, CQ3, CQ5, CQ6.
The temporal information of a \texttt{MusicAnnotation} and a \texttt{MusicObservation} is expressed in the same way, thus effectively creating an independent pattern for describing musical time intervals.
This pattern is composed by a \texttt{MusicTimeInterval} which in turn defines a \texttt{MusicTimeIndex} and a \texttt{MusicTimeDuration}. 
They indicate the time frame, within the referenced musical object, addressed by a music annotation/observation. 
More specifically, a \texttt{MusicTimeIndex} defines the start point of the annotation, while \texttt{MusicTimeDuration} describes the duration of the annotation.

%An annotation has a time validity (\texttt{hasValidityDuration}) expressed as a \texttt{xsd:float} value. Similarly, an \texttt{Observation} has a duration (\texttt{hasDuration}) expressed as a \texttt{xsd:float} value. They indicate the time frame, within the referenced musical object, addressed by an annotation/observation. These datatype properties can represent the duration of a score fragment (duration in beats) as well as the duration of the an audio fragment (duration in seconds). 

\noindent
Each \texttt{MusicTimeIndex} is composed of one or more components, namely \texttt{MusicTimeIndexComponent}s. The latter, as well as the \texttt{MusicTimeDuration}, defines the value of the temporal annotation via a datatype property \texttt{hasTimeValue}, which has as range \texttt{rdfs:Literal}, and the format of the annotation itself, expressed by the \texttt{MusicTimeValueType} class.
%The starting point of the time frame addressed by an annotation/observation is encoded differently depending on their type. For \texttt{AudioAnnotation} and \texttt{AudioObservation}, the starting time is defined as a \texttt{xsd:float} value (seconds) through the datatype properties \texttt{hasValidityStartingTime} and \texttt{startsAtTime}, respectively.

\noindent
In the case of \texttt{AudioMusicAnnotation} and \texttt{AudioMusicObservation}, the start time of the annotation shall be expressed by a single \texttt{MusicTimeIndexComponent}, which will have as \texttt{MusicTimeValueType} a time format in seconds, minutes or milliseconds. 
Instead, in the case of \texttt{ScoreAnnotation} and \texttt{ScoreObservation} two \texttt{MusicTimeIndexComponent}s will be needed to describe the start time, the first to describe the beat in which the annotation begins and the second to describe the beat within the measure in which the annotation starts.

\begin{lstlisting}
Class: MusicTimeInterval
    SubClassOf: 
        hasMusicTimeDuration only MusicTimeDuration,
        hasMusicTimeIndex only MusicTimeIndex,
        hasMusicTimeDuration exactly 1 MusicTimeDuration,
        hasMusicTimeIndex exactly 1 MusicTimeIndex
\end{lstlisting}

\begin{lstlisting}
Class: MusicTimeIndex
    SubClassOf: 
        hasMusicTimeIndexComponent only MusicTimeIndexComponent,
        hasMusicTimeIndexComponent min 1 MusicTimeIndexComponent
\end{lstlisting}

\begin{lstlisting}
Class: MusicTimeIndexComponent
    SubClassOf: 
        hasMusicTimeValueType only MusicTimeValueType,
        hasMusicTimeValueType exactly 1 MusicTimeValueType,
        hasTimeValue only rdfs:Literal,
        hasTimeValue exactly 1 rdfs:Literal
\end{lstlisting}

\paragraph{Annotator.} Addressing CQ8. Annotations have one and only one \texttt{Annotator}, relation expressed through the object property \texttt{hasAnnotator}. \texttt{Annotator}s are classified by their type (\texttt{AnnotatorType}), for example \texttt{Human}, \texttt{Machine}, \texttt{Crowdsourcing}, etc., which is exactly one.

\begin{lstlisting}
ObjectProperty: hasAnnotator
    SubPropertyChain: 
        isAnnotatorOf o includesMusicObservation
    Domain: 
        MusicAnnotation
    Range: 
        Annotator
\end{lstlisting}

\paragraph{Music Observation.} Addressing CQ1, CQ4, CQ7, CQ9.
Each \texttt{MusicAnnotation} includes a set of \texttt{MusicObservation}s. \texttt{MusicObservation}s can be of two types: \texttt{ScoreMusicObservation} and \texttt{AudioMusicObservation}. The type of an observation must be compatible with the type of the annotation that contains them.
If the annotation is \texttt{ScoreMusicAnnotation}, it contains \texttt{ScoreMusicObservation}s, otherwise it contains \texttt{AudioMusicObservation}s.
The annotator (and its type) of an observation is the same and only from the annotation that includes it: this is formalised by means of a property chain in the ODP. However, the level of confidence of an annotator is associated to each observation (\texttt{hasConfidence}).

\noindent
Each \texttt{MusicObservation} has an \texttt{MusicObservationValue}, which characterises its content. The \texttt{MusicObservationValue} class is meant to be specialised depending on the subject being observed (and annotated), e.g. Chord, Note, Structural Annotation. 
For example, it can generalise over concepts from existing ontologies, such as the Chord Ontology\footnote{Chord Ontology documentation available at: \url{http://motools.sourceforge.net/chord_draft_1/chord.html}} for chord annotations. 
%In fact, we believe that this pattern is quite flexible and can be adapted to a large number of annotations in the musical field.
\noindent
Musical object, music annotation, music observation, music observation value, music time interval, annotator, and annotator type are disjoint concepts.

\section{Usage Example}\label{sec:usage}

In this section, we describe two examples of usage of the \emph{Music Annotation} ODP. 
We remind that this ODP addresses different types of annotations for different types of sources (e.g. score, audio). The examples show how the Music Annotation ODP can be used to describe:
\begin{enumerate*}[label=(\roman*)]
  \item musical chord annotations and
  \item structural annotations of a song.
\end{enumerate*}

\begin{figure*}[h!]
    \centering
    \includegraphics[width=\textwidth]{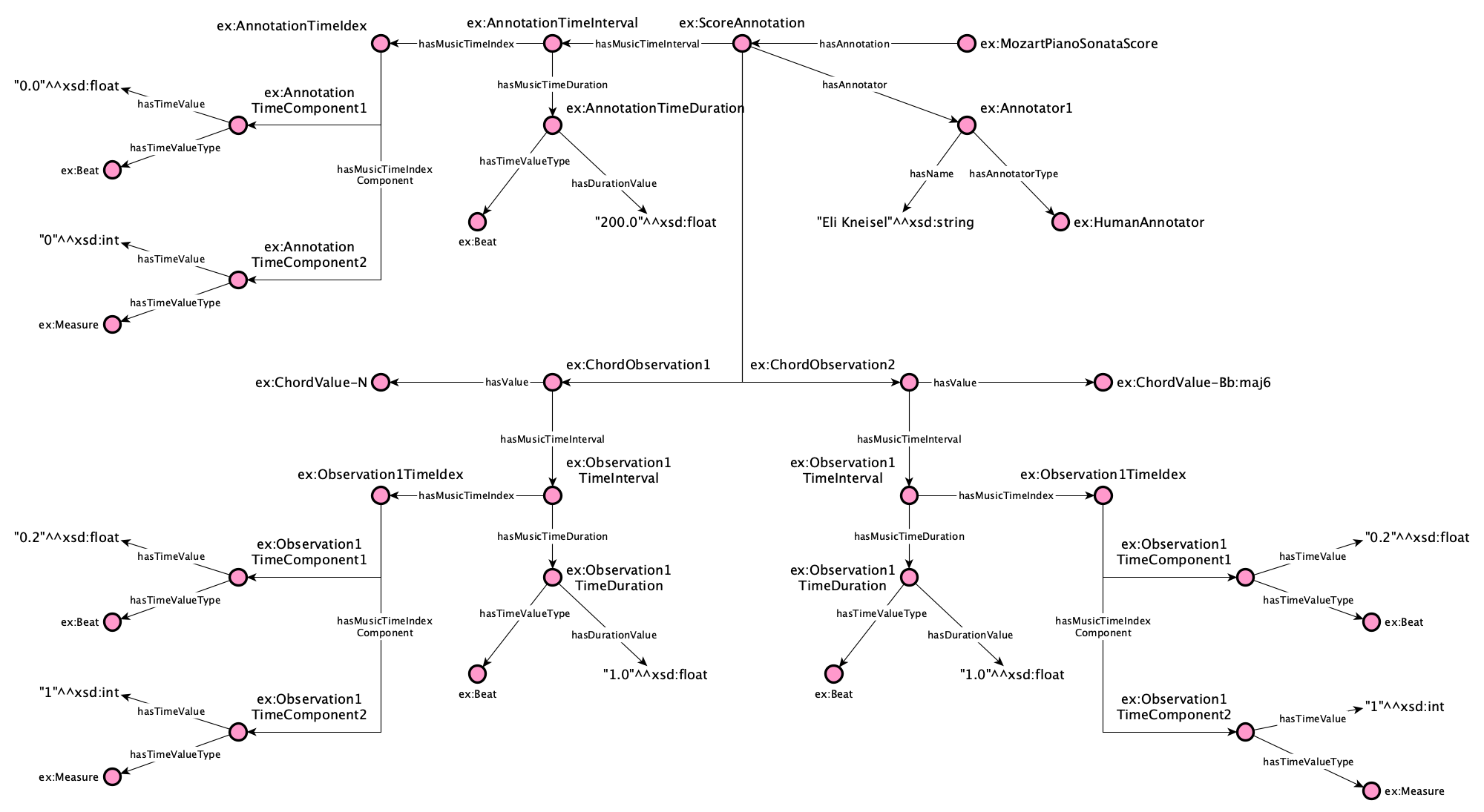}
    \caption{The Music Annotation ODP used to represent chord annotations from Wolfgang Amadeus Mozart's \emph{Piano Sonata no. 1 in C major (Allegro)}. The original annotation is taken from \cite{hentschel2021:mozartsonatas}.}
    \label{fig:annotation_chords}
\end{figure*}

\subsection{Chord Annotations}
The first example is an annotation of chords from a music score of Wolfgang Amadeus Mozart's \emph{Piano Sonata no. 1 in C major (Allegro)}. The original annotation is taken from the Mozart Piano Sonatas Dataset \cite{hentschel2021:mozartsonatas}.
%The original annotation is available in JAMS format and comes from the Isophonics\footnote{Isophonics dataset: \url{http://isophonics.net/datasets}} \cite{mauch09omras}.
Figure \ref{fig:annotation_chords} depicts the resulting RDF graph using the Grafoo Notation\footnote{\url{https://essepuntato.it/graffoo/}}. In all the examples, dummy prefix and namespace (\texttt{ex:} and \texttt{\url{http://example.org/}}) are defined for instances.

In this case, the \texttt{MusicalObject} is a musical score, defined by the \texttt{ex:MozartPianoSonataScore} instance, which has \texttt{ex:ScoreAnnotation} as its annotation.
The annotation is linked to its annotator, in this case a human and to its \texttt{MusicTimeInterval}. 
The \texttt{MusicTimeInterval} defines the duration of the annotation, by means of the \texttt{MusicTimeDuration} class, and the start point of the annotation, by means of the \texttt{MusicTimeIndex} class.
The latter, being the annotation is of type \emph{score}, contains two different \texttt{MusicTimeIndexComponent}s: the first has as its \texttt{MusicTimeValueType} a \texttt{ex:Measure}, which indicated the measure at which the annotation starts, while the second has as value type a \texttt{ex:Beat}, which describes the beat within the measure at which the annotation begins. Duration is instead expressed only in \emph{beats}. 

%Being an annotation of type \emph{score}, the starting time is expressed by means of the instance \texttt{ex:MozartScoreStartTime} that has defined beat and measure, via a \texttt{xsd:float} and an \texttt{xsd:int}, respectively.

The annotation then contains two different observations (the actual number has been reduced for demonstration purposes), namely \texttt{ex:ChordObservation1} and \texttt{ex:ChordObservation2}.

Each of these observations has a value, i.e. the chord per se, and a time interval. 
%Similarly to the annotation, the time information is expressed with a \texttt{MusicTimeInterval} class. 
In this example, observations have no \texttt{Confidence}, as this is not provided by the original annotation.

\subsection{Structural Annotations}

\begin{figure*}[h!]
    \centering
    \includegraphics[width=\textwidth]{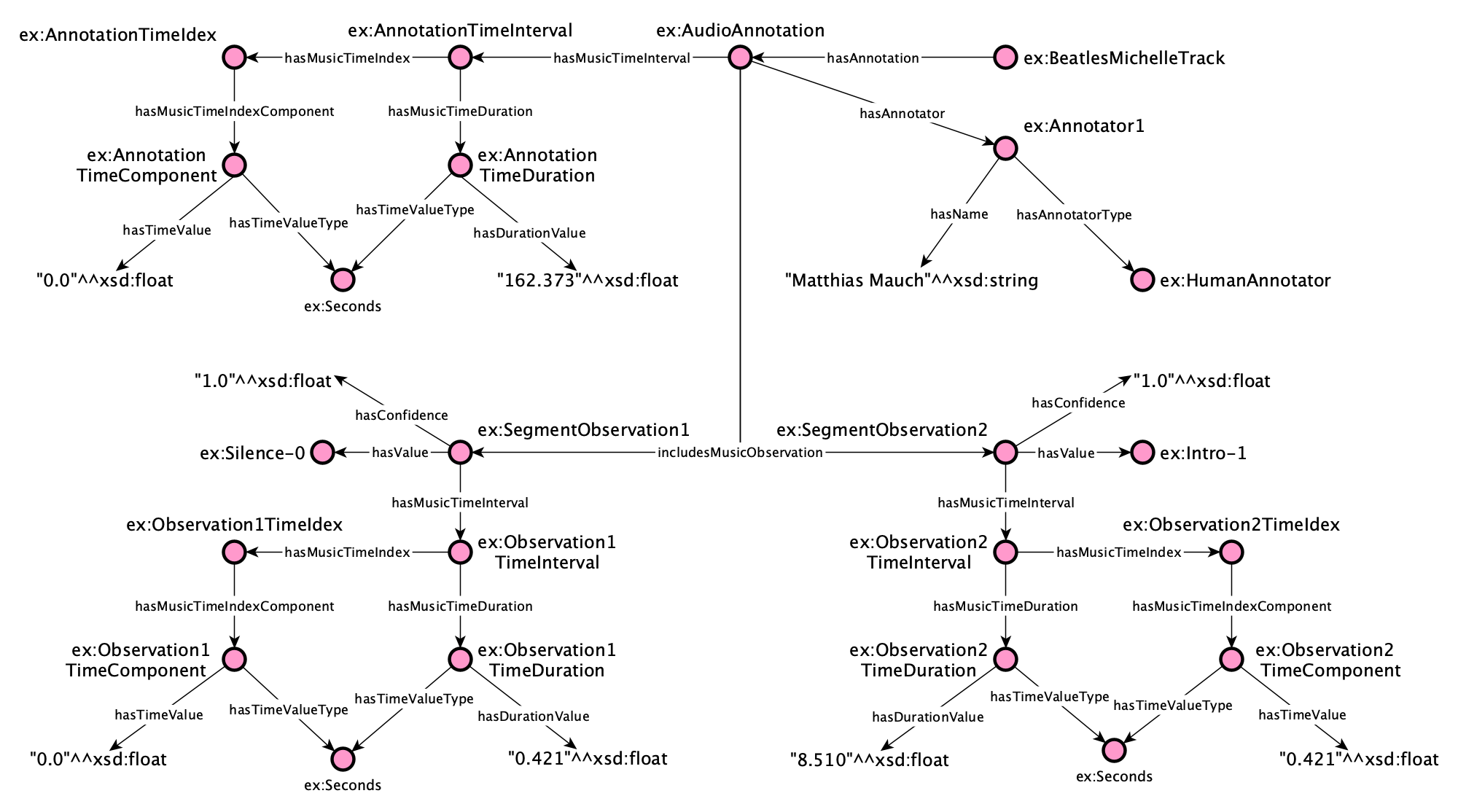}
    \caption{The Music Annotation ODP used to represent segment annotations from an audio track of The Beatles' \emph{Michelle}. The original annotation is taken from the Isophonics.}
    \label{fig:annotation_segments}
\end{figure*}

The second example is an annotation of segments from an audio track of The Beatles' \emph{Michelle}. The original annotation is available in JAMS format and is taken from the Isophonics\footnote{Isophonics dataset: \url{http://isophonics.net/datasets}} \cite{mauch09omras}.
Figure \ref{fig:annotation_segments} depicts the example graphically using the Grafoo notation. 

In this example, the \texttt{MusicalObject} is instead a track, defined by the \texttt{ex:BeatlesMichelleTrack} instance, which has an \texttt{ex:AudioMusicAnnotation}, as it was annotated from the audio signal. 
The annotation has a human-type annotator and an annotation time interval.

The annotation then contains two different \texttt{SegmentObservation}, which define the structure of the track. 
Each observation has a starting time and duration, defined by the classes \texttt{MusicTimeIndex} and \texttt{MusicTimeDuration}, respectively. In this case, there is only a single \texttt{MusicTimeIndexComponent}, since the time information is expressed in seconds (\texttt{ex:Seconds}). 
Finally, the value of each observation corresponds to the structural segment itself, in this case \texttt{ex:Silence} and \texttt{ex:Intro}.

\section{Conclusions and Future Work}\label{sec:conclusions}

We propose the \emph{Music Annotation} ODP for modelling annotations of music scores and audio tracks. A distinction at the core of this ODP is the different encoding of time information, which depends on the type of the subject of observation (score or audio). The ODP is the result of the analysis of many relevant different existing formats used for music annotation (MusicXML, ABC, JAMS, etc.) and provides a template for supporting the integration of data from such heterogeneous sources.
This work demonstrated the use of the ODP for modelling harmonic and structural annotations (chords, segments) collected from symbolic and audio sources.
We plan to follow up with a large scale integration experiment on a selection of MIR datasets, and the extension of our pattern to model additional types of music annotations.

%The need for this pattern emerges from the lack of tools to describe musical annotations while preserving the semantics of the annotations. 
% Our ODP is motivated by the need of integrating music annotation datasets that use different notation systems and relate to multi-modal sources.

% We also plan to extend the pattern and adapt it to other types of annotations, such as ...

%\todo[inline]{Future work: using the pattern for creating a JAMS ontology and creating a large KG of music annotations. (ChoCo)}

\begin{acknowledgments}
This project has received funding from the European Union’s Horizon 2020 research
and innovation programme under grant agreement No 101004746.
\end{acknowledgments}

%%
%% Define the bibliography file to be used
\bibliography{bibliography}

%%
%% If your work has an appendix, this is the place to put it.
\appendix

\end{document}